\documentclass{article}
\usepackage{ijcai24}

\usepackage{times}
\usepackage{soul}
\usepackage{url}
\usepackage[hidelinks]{hyperref}
\usepackage[utf8]{inputenc}
\usepackage[small]{caption}
\usepackage{graphicx}
\usepackage{amsmath}
\usepackage{amsthm}
\usepackage{booktabs}
\usepackage{algorithm}
\usepackage{algorithmic}
\usepackage[switch]{lineno}
\usepackage{caption}

\usepackage{color,xcolor}
\usepackage{colortbl}

\usepackage{pifont}

\usepackage[font=smaller]{caption}
\usepackage{multirow}
\usepackage{subcaption} %

\usepackage{amsmath,amsfonts,amssymb}
\usepackage{bm}
\usepackage{dsfont}

\newcommand{\myPara}[1]{\vspace{.04in}\noindent\textbf{#1}}
\newcommand{\dsyn}[1]{\mathcal{D}_\mathsf{syn}}
\newcommand{\dreal}[1]{\mathcal{D}_\mathsf{real}}
\DeclareMathOperator*{\argmin}{arg\,min}
\newcommand{\s}[1]{$_{\pm#1}$}

\title{DANCE: Dual-View Distribution Alignment for Dataset Condensation}

\author{
Hansong Zhang$^{1,2}$
\and
Shikun Li$^{1,2}$\and
Fanzhao Lin$^{1,2}$\and
Weiping Wang$^{1,2}$\\
Zhenxing Qian$^3$\and
Shiming Ge$^{1,2}$\thanks{Shiming Ge is the corresponding author.}\\
\affiliations
$^1$Institute of Information Engineering, Chinese Academy of Sciences, Beijing 100092, China\\
$^2$School of Cyber Security, University of Chinese Academy of Sciences, Beijing 100049, China\\
$^3$School of Computer Science, Fudan University, Shanghai 200433, China\\
\emails
\{zhanghansong,lishikun,linfanzhao,wangweiping,geshiming\}@iie.ac.cn,
zxqian@fudan.edu.cn
}

\begin{document}
\maketitle

\begin{abstract}
    Dataset condensation addresses the problem of data burden by learning a small synthetic training set that preserves essential knowledge from the larger real training set. To date, the state-of-the-art (SOTA) results are often yielded by optimization-oriented methods, but their inefficiency hinders their application to realistic datasets. On the other hand, the Distribution-Matching (DM) methods show remarkable efficiency but sub-optimal results compared to optimization-oriented methods. In this paper, we reveal the limitations of current DM-based methods from the inner-class and inter-class views, \textit{i.e.}, \textit{Persistent Training} and \textit{Distribution Shift}. To address these problems, we propose a new DM-based method named \underline{\textbf{D}}ual-view distribution \underline{\textbf{A}}lig\underline{\textbf{N}}ment for dataset \underline{\textbf{C}}ond\underline{\textbf{E}}nsation (DANCE), which exploits a few pre-trained models to improve DM from both inner-class and inter-class views. Specifically, from the inner-class view, we construct multiple ``middle encoders'' to perform pseudo long-term distribution alignment, making the condensed set a good proxy of the real one during the whole training process; while from the inter-class view, we use the expert models to perform distribution calibration, ensuring the synthetic data remains in the real class region during condensing. Experiments demonstrate the proposed method achieves a SOTA performance while maintaining comparable efficiency with the original DM across various scenarios. Source codes are available at \url{https://github.com/Hansong-Zhang/DANCE}.
    \vspace{-5pt}
\end{abstract}

\section{Introduction}
Recently, the reliance on large-scale datasets, which may include millions or even billions of examples, has become essential for developing state-of-the-art (SOTA) models~\cite{dsa,moderate_core,9661404,Li2023TAIDTM,zhang2024ccc}. However, this reliance brings significant challenges, primarily due to the substantial storage costs and computational expenses required for training such models. These challenges pose formidable obstacles, particularly for startups and non-profit organizations, making advanced model training often unattainable~\cite{highcost2,highcost3,zheng2022coverage,jin2022condensing,yang2022dataset,geng2023survey,Xia2024RefinedCS}.

\begin{figure}[t]
    \centering
        \includegraphics[width=0.49\textwidth]{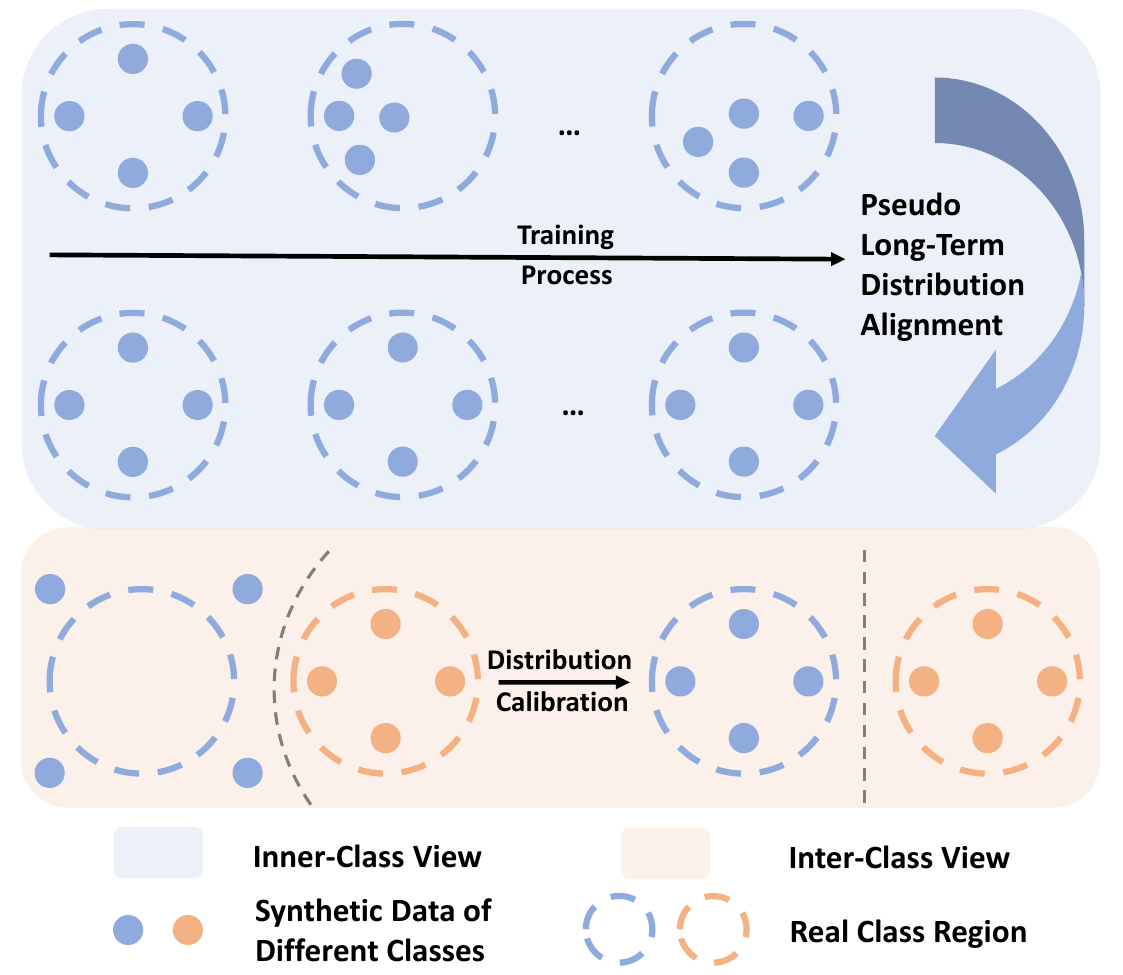}
    \caption{\textbf{Two views of the proposed DANCE.} For inner-class view, it ensures that the synthetic data remains a faithful proxy of the real data throughout the training process. For inter-class view, it also prevents the synthetic data from falling outside the real class region (the domain where all real data points of that class reside), which may change the decision boundary of the learned classifier.}
    \vspace{-10pt}
    \label{fig:first_page}
\end{figure}

As a remedy, \textit{Dataset Condensation} (DC), also known as \textit{Dataset Distillation}, has emerged as a prominent solution to address the challenges of data burden~\cite{dd_wang,cui2022dc,yu2023dataset}. It involves learning a small condensed training set to replicate the performance of models trained on larger real datasets. Pioneer methods in this area typically focus on matching either the gradients~\cite{dsa,dc,idc,rev} or parameters~\cite{mtt,du2023minimizing,guo2023datm,liu2022dataset} between real and synthetic data, which can be categorized as \textit{Optimization-Oriented} methods. While these methods have shown success, their reliance on the bi-level optimization or nested gradients often results in prohibitively high computational costs~\cite{modelaugmentation,sajedi2023datadam,liu2021investigating,zhang2024m3d}, limiting their practical application in wider scenarios.

To address the scalability challenges in DC, \textit{Distribution Matching} (DM)~\cite{dm} has been proposed. It focuses on aligning the latent representations extracted by randomly-initialized encoders, based on the rationale that the condensed set should represents the real training set in the feature space. Unlike previous \textit{Optimization-Oriented} methods, DM avoids the computationally intensive nested optimization loops, significantly reducing the time required for condensation and thereby enhancing its applicability in diverse scenarios~\cite{loo2022efficient,zhou2023dataset,cazenavette2023generalizing,liu2023slimmable,KIP}. However, despite these advantages, DM's performance still falls short of SOTA optimization-oriented methods such as MTT~\cite{mtt}, IDC~\cite{idc}, and DREAM~\cite{liu2023dream}.

In this paper, we conduct an in-depth analysis of DM from the inner-class and inter-class views, pointing out the limitations of current DM-based methods and provide our corresponding remedies. Specifically, from the \textbf{inner-class view}, to ensure an alignment during the whole training process, previous works like IDM~\cite{idm} and CAFE~\cite{cafe} naively use the models trained from scratch to extract the latent representations. While effective, the \textit{Persistent Training}, i.e. numerous model updating steps, is very time-consuming thus greatly hinders their efficiency. To counter this, we introduce Pseudo Long-Term Distribution Alignment (PLTDA), where we use the convex combination of initialized and trained expert models to perform inner-class distribution alignment, eliminating the need for persistent training. From the \textbf{inter-class view}, we reveal the \textit{Distribution Shift} phenomenon in DM, i.e., the synthetic data will diverge from the real class region during condensation, which may change the decision boundary of the learned classifier. To address this, we employ expert models for Distribution Calibration, ensuring the synthetic data remains within the real class region. We term the proposed method as \underline{\textbf{D}}ual-view distribution \underline{\textbf{A}}lignme\underline{\textbf{N}}t for dataset \underline{\textbf{C}}ond\underline{\textbf{E}}nsation (DANCE), for we enhance DM by utilizing the knowledge of expert models from the above two views, which is illustrated in Fig.~\ref{fig:first_page}.
As will be shown in the experiments, DANCE can achieve comparable results to SOTA optimization-oriented methods even with only a single expert model.

Our main contributions are outlined as follows:

\textbf{[C1]}: We identify and analyze the limitations of current DM-based dataset condensation methods from inner- and inter-class views, which reveals two major issues: persistent training and distribution shift.

\textbf{[C2]}: We introduce DANCE by incorporating two modules to effectively mitigate the above two issues inherent in DM-based methods.

\textbf{[C3]}: We conduct extensive experiments across a variety of datasets under different resolutions. The results demonstrate that DANCE establishes a strong baseline in dataset condensation, significantly advancing both performance and efficiency, particularly in the realm of distribution matching.

\section{Preliminaries}
In this section, we initially formalize the concept of dataset condensation (DC) and then recap the DM method~\cite{dm}, which is pivotal as it represents the pioneering work in the realm of distribution matching within DC and lays the groundwork for our research.

\paragraph{Problem Definition.}
Given a large real training set $\dreal{}=\{(\bm{x}^{\mathsf{real}}_i, y^{\mathsf{real}}_i)\}_{i=1}^{|\dreal{}|}$,
\textit{Dataset Condensation} or \textit{Dataset Distillation} aims to generate a small training set $\dsyn{}=\{(\bm{x}^{\mathsf{syn}}_j, y^{\mathsf{syn}}_j)\}_{j=1}^{|\dsyn{}|}$ ($|\dsyn{}|\ll |\dreal{}|$), so that the model trained on $\dsyn{}$ and the model trained on $\dreal{}$ (denoted as $\bm{\theta}_\mathsf{syn}$ and $\bm{\theta}_\mathsf{real}$ respectively) will have similar performance on the unseen data. Formally, let $P_D$ represent the distribution of the real data, $\ell$ be the loss operation such as cross-entropy, the synthetic training set can be obtained by minimizing the performance gap between the two models:
\begin{equation}
    \dsyn{}^\star = \argmin_{\mathbb{E}_{(\bm{x},y)\sim P_D}}||\ell(\bm{\theta}_\mathsf{syn}(\bm{x}),y) - \ell(\bm{\theta}_\mathsf{real}(\bm{x}),y)||.
    \label{dc_concept}
\end{equation}

\paragraph{Distribution Matching.}
To solve Eq.~(\ref{dc_concept}), previous optimization-oriented methods have attempted to 1) update the $\dsyn{}$ using a meta-learning framework. 2) match the gradient or parameter induced by $\dsyn{}$ and $\dreal{}$. However, both the above methods involves a bi-level optimization, which is computationally inefficient due to the calculation of nested gradients. To improve the condensing efficiency, DM~\cite{dm} first proposed Distribution Matching, which learns the condensed set by aligning the feature distributions of $\dsyn{}$ and $\dreal{}$. Specifically, the condensed set in DM is optimized by:
\begin{equation}
    \dsyn{}^\star = \argmin_{\mathbb{E}_{\bm{\phi}_0\sim P_{\bm{\phi}_0}}}\left\|\frac{\sum_{i=1}^{|\dreal{}|}\bm{\phi}_0(\bm{x}_i^\mathsf{real})}{|\dreal{}|} - \frac{\sum_{j=1}^{|\dsyn{}|}\bm{\phi}_0(\bm{x}_j^\mathsf{syn})}{|\dsyn{}|}\right\|^2,
    \label{dm_target}
\end{equation}
where $\bm{\phi}_0 \sim P_{\bm{\phi}_0}$ denotes the randomly-initialized feature extractor (instanced by a random DNN $\bm{\theta}_0$ without the linear classification layer).
Compared to optimization-oriented methods, DM significantly enhances the computational efficiency and has shown better generalization ability across different architectures~\cite{dm}.

\begin{figure*}[t]
    \centering
    \includegraphics[width=1.0\textwidth]{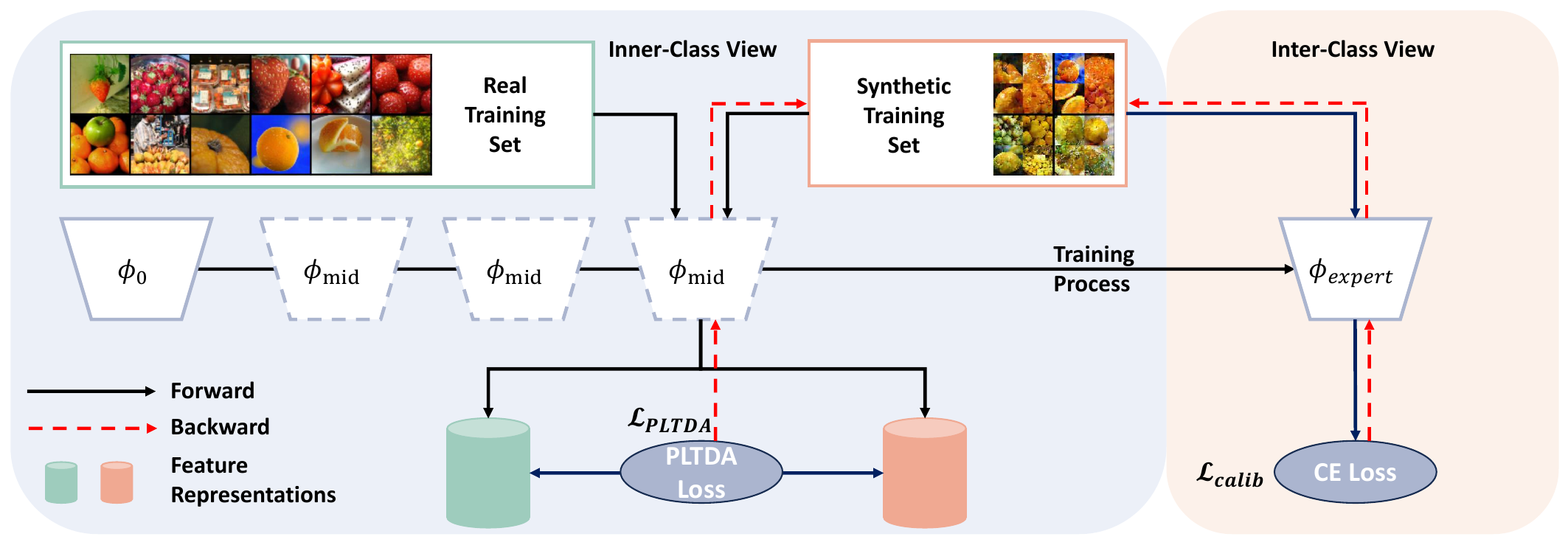}
    \caption{\textbf{The framework of DANCE}. From the inner-class view, multiple middle encoders are constructed to perform Pseudo Long-Term Distribution Alignment so that the synthetic set can remain a good proxy of its class during training. From the inter-class view, the Distribution Calibration is performed, ensuring the synthetic data stay within the real class region during condensing process.}
    \label{fig:framework}
\end{figure*}

\section{Methodology}

While DM~\cite{dm} brings remarkable efficiency and cross-architecture performance, the quality of the condensed set it generates typically falls short of those produced by SOTA optimization-oriented methods like IDC~\cite{idc} and MTT~\cite{mtt}. In this paper, we aim to enhance the alignment between the distributions of $\dreal{}$ and $\dsyn{}$, considering both the \textbf{inner-class view} and the \textbf{inter-class view}. Sections~\ref{sec:inner} and \ref{sec:inter} will detail the limitations of current DM-based methods from these two perspectives and introduce our proposed solutions. Subsequently, we describe our overall training algorithm in Section~\ref{sec:train_algor}. Our method, termed \underline{\textbf{D}}ual-view distribution \underline{\textbf{A}}lig\underline{\textbf{N}}ment for dataset \underline{\textbf{C}}ond\underline{\textbf{E}}nsation (DANCE), is depicted in Fig.~\ref{fig:framework}.

\subsection{Inner-Class View}\label{sec:inner}
\begin{figure*}[tb]
    \centering
    \begin{subfigure}[t]{0.33\textwidth}
        \centering
        \includegraphics[width=\textwidth]{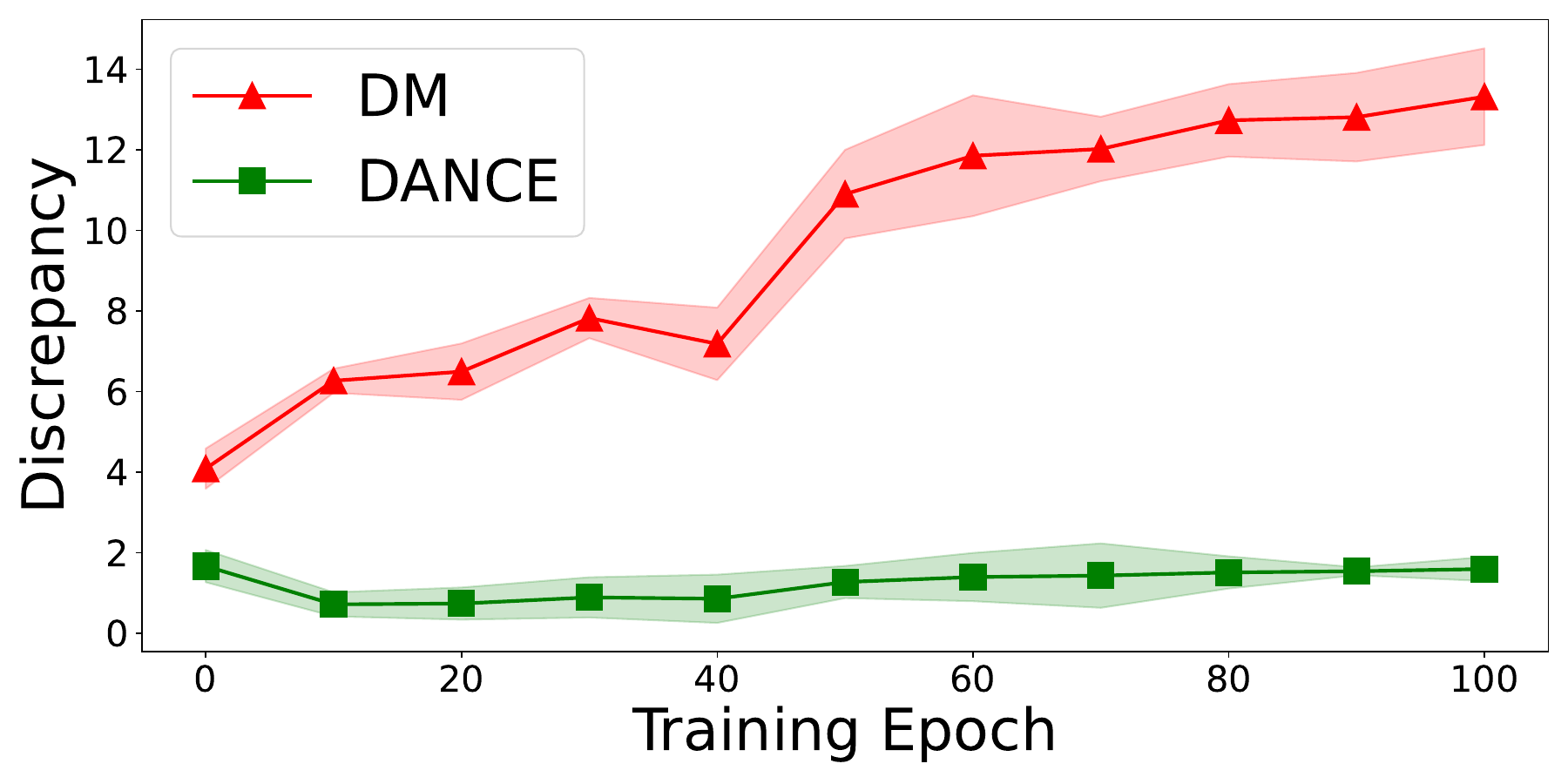}
        \caption{}
        \label{fig:misalignment}
    \end{subfigure}
    \hfill
    \begin{subfigure}[t]{0.33\textwidth}
        \centering
        \includegraphics[width=\textwidth]{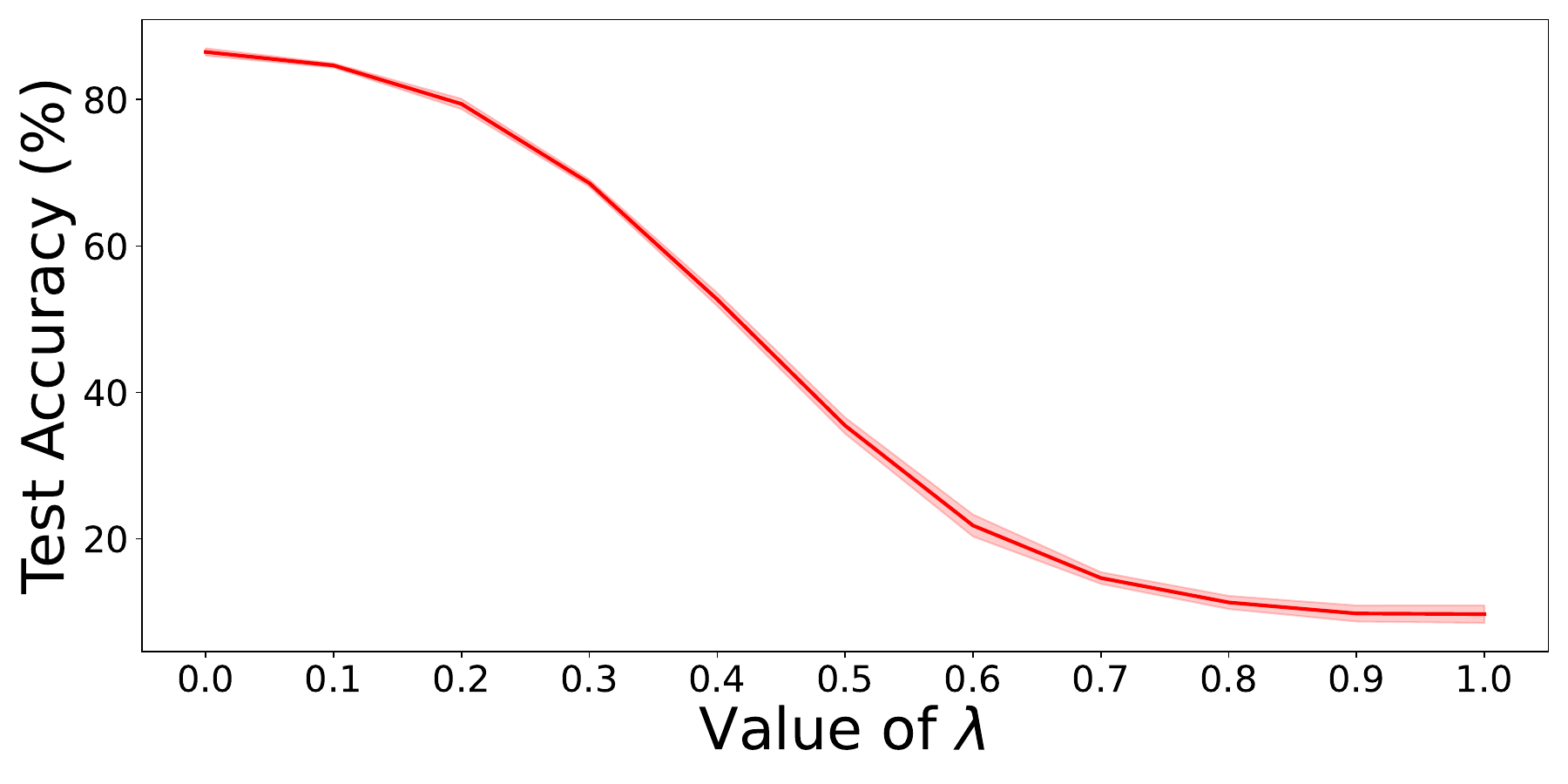}
        \caption{}
        \label{fig:different_stage}
    \end{subfigure}
    \hfill
    \begin{subfigure}[t]{0.33\textwidth}
        \centering
        \includegraphics[width=\textwidth]{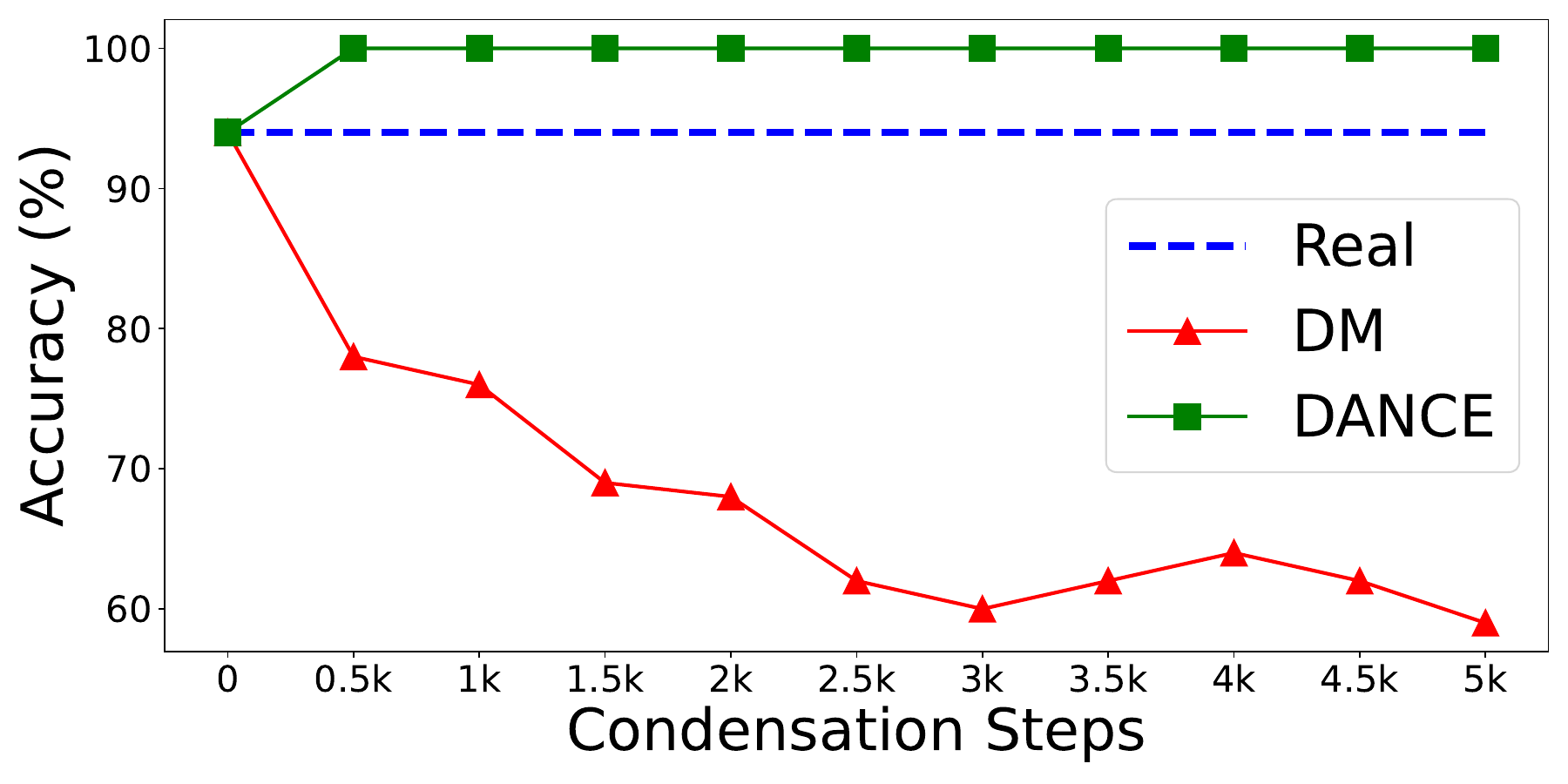}
        \caption{}
        \label{fig:motive_inter}
    \end{subfigure}
    \caption{(a) \textbf{The discrepancy between the feature distribution of $\dreal{}$ and $\dsyn{}$ of DM and DANCE across the whole training process with the real training data.} (b) \textbf{The test accuracy(\%) of the middle model $\bm{\phi}_\mathsf{mid}$ at different value of $\lambda$.} (c) The \textbf{accuracy (\%) of the expert model on the real test data, and the synthesized data of DM and DANCE.}
    The evaluations are conducted on CIFAR-10, where (a) and (c) adopts 10 images per class.}
    \label{fig:motive}
\end{figure*}

\paragraph{Limitation of DM.}
For data from the same class, DM~\cite{dm} employs various randomly-initialized deep encoders to extract latent representations of $\dreal{}$ and $\dsyn{}$. It then minimizes the discrepancy of feature distributions to ensure they are aligned (Eq.~(\ref{dm_target})). However, we contend that aligning feature distributions from randomly initialized extractors, which are sampled from a probability distribution over parameters $P_{\bm{\phi}_0}$, does not ensure that $\dsyn{}$ remains a reliable proxy for $\dreal{}$ throughout all training stages.
This divergence may ultimately cause the model trained on $\dsyn{}$ to deviate from the one trained on $\dreal{}$. To illustrate this divergence throughout the entire training process by the real training data, we compute the discrepancy between the distribution of $\dsyn{}$ and $\dreal{}$ at various stages of training. As depicted in Fig.~\ref{fig:misalignment}, DM fails to maintain the informativeness of the condensed set throughout the training procedure. During training, the distribution of the data condensed by DM increasingly deviates from that of the real data.

\paragraph{\textit{Remark}.} The misalignment issue is also noted in CAFE~\cite{cafe} and IDM~\cite{idm}. As a remedy, these approaches involve training multiple models from scratch to extract features of $\dsyn{}$ and $\dreal{}$ during the condensation process. While such ~\textit{Persistent Training} yields improved performance, it is hampered by the following two drawbacks:

\paragraph{Drawbacks.} 1) \textit{Hyper-parameter tuning}: The models used for condensation need to be \textbf{persistently} trained during the condensing process in CAFE and IDM to enrich the distilled knowledge. This process, however, involves meticulous tuning of multiple parameters, including the number of training steps, iteration count, and learning rate during the model-updating phase. 
Moreover, as model performance often sees a significant rise at the onset of training, the models at early stages are prone to be skipped due to overdoing the update steps, thereby hampering the effectiveness of the condensed images. 
2) \textit{Inefficiency}: Both CAFE and IDM necessitate optimizing hundreds of random models each time a dataset is condensed, which is impractical, especially for datasets with larger resolution.

\paragraph{Pseudo Long-Term Distribution Alignment (PLTDA).}
To tackle the misalignment issue from the inner-class perspective, we introduce a straightforward and effective module, termed \textbf{P}seudo \textbf{L}ong-\textbf{T}erm \textbf{D}istribution \textbf{A}lignment (PLTDA). Specifically, rather than relying on models trained on real data, we employ a convex combination of randomly-initialized encoders and their corresponding trained counterparts. We refer to these trained encoders as ``expert encoders'' ($\bm{\phi}_\mathsf{expert}$), because their corresponding ``expert models'' ($\bm{\theta}_\mathsf{expert}$) represent the upper bound of the performance of $\dsyn{}$ and the end of the training process. We term this combination as ``middle encoders'' ($\bm{\phi}_\mathsf{mid}$) of ``middle models'' ($\bm{\theta}_\mathsf{mid}$), which is calculated by:
\begin{equation}
\label{eq:media_enc}
\begin{matrix}
    \bm{\phi}_0 & \rightarrow & \lambda \cdot \bm{\phi}_0 + (1-\lambda) \cdot \bm{\phi}_\mathsf{expert} & \rightarrow & \bm{\phi}_\mathsf{expert} \\
                                                                 &        &      \rotatebox[origin=c]{90}{=} & &\\
                      &           &  \bm{\phi}_\mathsf{mid} &                &    
\end{matrix}
\end{equation}
where $\lambda \sim U(0,1)$ is a randomly generated coefficient for encoder combination. After obtaining the middle encoder, we calculate the loss in PLTDA as:
\begin{equation}
\label{eq:LTDA_loss}
    \mathcal{L}_\mathsf{PLTDA} = \left\|  \frac{\sum_{i=1}^{|\dreal{}|}\bm{\phi}_\mathsf{mid}(\bm{x}^\mathsf{real}_i)}{|\dreal{}|} - \frac{\sum_{j=1}^{|\dsyn{}|}\bm{\phi}_\mathsf{mid}(\bm{x}^\mathsf{syn}_j) }{|\dsyn{}|} \right\|^2.
\end{equation}

Compared to CAFE and IDM, which update the model during the condensation process, our expert encoders do not introduce additional hyper-parameters. Furthermore, as illustrated in Fig.~\ref{fig:different_stage}, the performance of the middle models changes smoothly across different values of $\lambda$, allowing for the generation of models with various performances. As shown in Fig.~\ref{fig:misalignment}, this way can make the distribution of the condensed data match the real data at all training stages.
Additionally, the middle encoders rely solely on the randomly-initialized encoder $\bm{\phi}_0$ and the expert encoders $\bm{\phi}_\mathsf{expert}$. Both can be pre-trained offline and reused for different values of images per class (IPCs). Notably, our method does not require a large number of pre-trained expert models like IDC~\cite{idc} or MTT~\cite{mtt}. As we will demonstrate in Sec.~\ref{sec:ablation}, our approach achieves state-of-the-art results across various scenarios with just a single expert model.

\subsection{Inter-Class View}\label{sec:inter}
\paragraph{Limitation of DM.} As depicted in~\cite{dm}, DM aligns the distribution between $\dsyn{}$ and $\dreal{}$ in a class-wise manner, while overlooking inter-class constraints.
Since many existing DM-based methods~\cite{sajedi2023datadam,zhang2024m3d} follow the class-wise learning manner of DM, these approaches may have the following drawback:

\paragraph{Drawback.}\textit{Distribution Shift}: 
The synthetic data may fall outside the real class region during condensing, which will affect the decision boundary of the learned classifier. As shown in Fig.\ref{fig:motive_inter}, the expert model trained by real training data has an excellent performance on real test data, while it cannot achieve a high classification accuracy on the synthetic data generated by DM. It can be inferred from this phenomenon that, many of the synthetic examples are not within the real class region and even cross the decision boundary of the expert model.

\paragraph{Distribution Calibration.} 
To address the above issue from the inter-class view, we integrate a module called \textit{Distribution Calibration} into our approach. This module utilizes expert models $\bm{\theta}_\mathsf{expert}$ to calibrate the inter-class distribution of $\dsyn{}$ after PLTDA. Specifically, once the inner-class matching is completed, we impose the following calibration loss, computed using the chosen $\bm{\theta}_\mathsf{expert}$, to prevent the synthetic data from straying from their respective categories:
\begin{equation}
\label{eq:calib_loss}
    \mathcal{L}_\mathsf{calib} = \frac{1}{|\dsyn{}|}\sum_{j=1}^{|\dsyn{}|}\ell(\bm{\theta}_\mathsf{expert}(\bm{x}^{\mathsf{syn}}_j), y^{\mathsf{syn}}_j).
\end{equation}
It is important to note that, although ``Discrimination Loss'' in CAFE~\cite{cafe} and ``Distribution Regularization'' in IDM~\cite{idm} employ a similar concept, they utilize un-converged models for calculating their losses, instead of expert models. This may lead to sub-optimal outcomes due to the relatively poorer generalization ability of such models compared to expert models.

\subsection{Training Algorithm}\label{sec:train_algor}
The pseudo-code of DANCE is detailed in Algorithm~\ref{alg:algorithm}. Besides incorporating the PLTDA (Sec.~\ref{sec:inner}) and Distribution Calibration (Sec.~\ref{sec:inter}), our approach also integrates a prevalent data augmentation technique known as ``Factoring \& Up-sampling''. In this technique, each image space in $\dsyn{}$ is divided into $l\times l$ smaller sections in order to host multiple synthetic images. These mini-images are subsequently up-sampled to their original dimensions during model training. This augmentation strategy was initially introduced by IDC~\cite{idc} and has since been widely employed in various dataset condensation works~\cite{liu2023dream,idm}.

\begin{algorithm}[tb]
    \caption{Dual-View Distribution Alignment for Dataset Condensation}
    \label{alg:algorithm}
    \textbf{Input}: Real training set $\dreal{}$\\
    \textbf{Parameter}: Number of expert models $N$; Number of condensation iterations $I$; Learning rate of the condensed set $\eta$; Calibration interval $I_c$\\
    \textbf{Output}: The condensed set $\dsyn{}$
    \begin{algorithmic}[1] 
        \STATE Initialize $\dsyn{}$ with randomly selected real data
        \STATE Pre-train $N$ expert encoders $\{\bm{\phi}_\mathsf{expert}^n\}_{n=1}^N$ and save their corresponding initial encoders $\{\bm{\phi}_0^n\}_{n=1}^N$
        \FOR{$i=1,2,\dots,I$}
        \STATE Randomly select an expert encoder $\bm{\phi}_\mathsf{expert}^n$ and generate the middle encoder $\bm{\phi}_\mathsf{mid}^n$ by Eq.~(\ref{eq:media_enc})
        \STATE Calculate the matching loss $\mathcal{L}_\mathsf{PLTDA}$ by Eq.~(\ref{eq:LTDA_loss})
        \STATE Update the $\dsyn{}$ by $\dsyn{} = \dsyn{} - \eta \nabla_{\dsyn{}}\mathcal{L}_\mathsf{PLTDA}$
        \IF{$i\%I_c=0$}
        \STATE Calculate the calibration loss $\mathcal{L}_\mathsf{calib}$ by Eq.~(\ref{eq:calib_loss})
        \STATE Update the $\dsyn{}$ by $\dsyn{} = \dsyn{} - \eta \nabla_{\dsyn{}}\mathcal{L}_\mathsf{calib}$
        \ENDIF
        \ENDFOR        
        \STATE \textbf{Return}: $\dsyn{}$
    \end{algorithmic}
\end{algorithm}

\begin{table*}[tb]
    \centering
    \small
    \setlength\tabcolsep{2.7pt}
    \begin{tabular}{l|ccc|ccc|ccc|ccc}
    \toprule
              &  \multicolumn{3}{c|}{Fashion-MNIST}  &  \multicolumn{3}{c|}{CIFAR-10} & \multicolumn{3}{c|}{CIFAR-100} & \multicolumn{3}{c}{TinyImageNet}\\
             \midrule
        Resolution & \multicolumn{3}{c|}{28 $\times$ 28} & \multicolumn{3}{c|}{32 $\times$ 32} & \multicolumn{3}{c|}{32 $\times$ 32} & \multicolumn{3}{c}{64 $\times$ 64}\\
        IPC    & 1 & 10  & 50 &  1  & 10  & 50   & 1 & 10 & 50  & 1  & 10 & 50 \\
        Ratio (\%)     &  0.017 & 0.17 & 0.83  & 0.02 & 0.2 & 1 & 0.02 & 0.2 & 1 & 0.2 & 2 & 10 \\
        \midrule
        Random      & 51.4\s{3.8} & 73.8\s{0.7} & 82.5\s{0.7} & 14.4\s{2.0} & 26.0\s{1.2} & 43.4\s{1.0} & 4.2\s{0.3} &14.6\s{0.5} & 30.0\s{0.4} & 1.4\s{0.1} & 5.0\s{0.2} & 15.0\s{0.4}\\
        Herding     & 67.0\s{1.9} & 71.1\s{0.7} & 71.9\s{0.8} & 21.5\s{1.2} & 31.6\s{0.7} & 40.4\s{0.6} & 8.4\s{0.3} & 17.3\s{0.3} & 33.7\s{0.5}  & 2.8\s{0.2} & 6.3\s{0.2} & 16.7\s{0.3}\\
        K-Center    & 66.9\s{1.8} & 54.7\s{1.5} & 68.3\s{0.8} & 21.5\s{1.3} & 14.7\s{0.9} & 27.0\s{1.4} & 8.3\s{0.3} & 7.1\s{0.2}& 30.5\s{0.3} & - & - & -\\
        \midrule
        DC      & 70.5\s{0.6} & 82.3\s{0.4} & 83.6\s{0.4} & 28.3\s{0.5} & 44.9\s{0.5} & 53.9\s{0.5} & 12.8\s{0.3} & 25.2\s{0.3} & - & 5.3\s{0.1} & 12.9\s{0.1} & 12.7\s{0.4}\\
        DSA     & 70.6\s{0.6} & 84.6\s{0.3} & \cellcolor[HTML]{DAE8FC}{\textbf{88.7\s{0.2}}} & 28.8\s{0.7} & 52.1\s{0.5} & 60.6\s{0.5} & 13.9\s{0.3} & 32.3\s{0.3} & 42.8\s{0.4}  & 5.7\s{0.1} & 16.3\s{0.2} & 5.1\s{0.2}\\
        IDC     & 81.0\s{0.2} & 86.0\s{0.3} & 86.2\s{0.2} & {\textbf{50.6\s{0.4}}} & 67.5\s{0.5} & 74.5\s{0.1} & - & 45.1\s{0.4} & - & - & - & -\\
        DREAM   & {\textbf{81.3\s{0.2}}} & \cellcolor[HTML]{DAE8FC}{\textbf{86.4\s{0.3}}} & 86.8\s{0.3} & \cellcolor[HTML]{DAE8FC}{\textbf{51.1\s{0.3}}} & {\textbf{69.4\s{0.4}}} & {\textbf{74.8\s{0.1}}} & \cellcolor[HTML]{DAE8FC}{\textbf{29.5\s{0.3}}} & {\textbf{46.8\s{0.7}}} & {\textbf{52.6\s{0.4}}} & 10.0\s{0.4} & - & \cellcolor[HTML]{DAE8FC}{\textbf{29.5\s{0.3}}}\\
        MTT     & - & - & - & 31.9\s{1.2} & 56.4\s{0.7} & 65.9\s{0.6} & 24.3\s{0.3} & 40.1\s{0.4} & 47.7\s{0.2}  & 6.2\s{0.4} & 17.3\s{0.2} & 26.5\s{0.3}\\
        \midrule
        CAFE    & 77.1\s{0.9} & 83.0\s{0.4} & 84.8\s{0.4} & 30.3\s{1.1} & 46.3\s{0.6} & 55.5\s{0.6} & 12.9\s{0.3} & 27.8\s{0.3} & 37.9\s{0.3} & - & - & -\\
        CAFE+DSA & 73.7\s{0.7} & 83.0\s{0.3} & {\textbf{88.2\s{0.3}}} & 31.6\s{0.8} & 50.9\s{0.5} & 62.3\s{0.4} & 14.0\s{0.3} & 31.5\s{0.2} & 42.9\s{0.2} & - & - & -\\
        DM     & 70.7\s{0.6} & 83.5\s{0.3} & 88.1\s{0.6} & 26.0\s{0.8} & 48.9\s{0.6} & 63.0\s{0.4} & 11.4\s{0.3} & 29.7\s{0.3} & 43.6\s{0.4}  & 3.9\s{0.2} & 12.9\s{0.4} & 24.1\s{0.3}\\
        IDM    & - & - & - & 45.6\s{0.7} & 58.6\s{0.1} & 67.5\s{0.1} & 20.1\s{0.3} & 45.1\s{0.1} & 50.0\s{0.2} & {\textbf{10.1\s{0.2}}} & {\textbf{21.9\s{0.2}}} & 27.7\s{0.3}\\
        DataDAM & - & - & - & 32.0\s{1.2} & 54.2\s{0.8} & 67.0\s{0.4} & 14.5\s{0.5} & 34.8\s{0.5} & 49.4\s{0.3} & 8.3\s{0.4} & 18.7\s{0.3} & 28.7\s{0.3}\\
        \midrule
        \textbf{DANCE}   & \cellcolor[HTML]{DAE8FC}{\textbf{81.5\s{0.4}}} & {\textbf{86.3\s{0.2}}} & 86.9\s{0.1} & 47.1\s{0.2} & \cellcolor[HTML]{DAE8FC}{\textbf{70.8\s{0.2}}} & \cellcolor[HTML]{DAE8FC}{\textbf{76.1\s{0.1}}} & {\textbf{27.9\s{0.2}}} & \cellcolor[HTML]{DAE8FC}{\textbf{49.8\s{0.1}}} & \cellcolor[HTML]{DAE8FC}{\textbf{52.8\s{0.1}}}  & \cellcolor[HTML]{DAE8FC}{\textbf{11.6\s{0.2}}} & \cellcolor[HTML]{DAE8FC}{\textbf{26.4\s{0.3}}} & {\textbf{28.9\s{0.4}}}\\
        \midrule
        Whole Dataset & \multicolumn{3}{c|}{93.5\s{0.1}} & \multicolumn{3}{c|}{84.8\s{0.1}} & \multicolumn{3}{c|}{56.2\s{0.3}} & \multicolumn{3}{c}{37.6\s{0.4}} \\
    \bottomrule
    \end{tabular}
    \caption{\textbf{Comparison with previous coreset selection and dataset condensation methods on low-resolution datasets and medium-resolution datasets.} IPC: image(s) per class. Ratio $(\%)$: the ratio of condensed examples to the whole training set. Best results are \colorbox[HTML]{DAE8FC}{\textbf{highlighted}} and the second best results are in {\textbf{bold}}. Note that some entries are marked as ``-'' because of scalability issues or the results are not reported.}
    \label{tab:res_lowresolution}
\end{table*}

\begin{table*}[tb]
    \centering
    \small
    \setlength\tabcolsep{2.7pt}
    \begin{tabular}{l|cc|cc|cc|cc|cc|cc}
    \toprule
             &  \multicolumn{2}{c|}{ImageNette}  &  \multicolumn{2}{c|}{ImageWoof}  &  \multicolumn{2}{c|}{ImageFruit} &  \multicolumn{2}{c|}{ImageMeow} &  \multicolumn{2}{c|}{ImageSquawk} &  \multicolumn{2}{c}{ImageYellow} \\
             \midrule
        IPC    & 1 & 10 & 1 & 10 & 1 & 10 & 1 & 10 & 1 & 10 & 1 & 10 \\
        Ratio (\%)     & 0.105 & 1.050 & 0.110 &  1.100 & 0.077 & 0.77  & 0.077 & 0.77  & 0.077 & 0.77 & 0.077 & 0.77 \\
        \midrule
        Random     & 23.5\s{4.8} & 47.7\s{2.4} & 14.2\s{0.9} & 27.0\s{1.9} & 13.2\s{0.8} & 21.4\s{1.2} & 13.8\s{0.6} & 29.0\s{1.1} & 21.8\s{0.5} & 40.2\s{0.4} & 20.4\s{0.6} & 37.4\s{0.5} \\
        MTT     & {\textbf{47.7\s{0.9}}} & {\textbf{63.0\s{1.3}}} & {\textbf{28.6\s{0.8}}} & {\textbf{35.8\s{1.8}}} & {\textbf{26.6\s{0.8}}} & {\textbf{40.3\s{1.3}}} & {\textbf{30.7\s{1.6}}} & {\textbf{40.4\s{2.2}}} & {\textbf{39.4\s{1.5}}} & 52.3\s{1.0} & {\textbf{45.2\s{0.8}}} & {\textbf{60.0\s{1.5}}} \\
        DM     & 32.8\s{0.5} & 58.1\s{0.3} & 21.1\s{1.2} & 31.4\s{0.5} & - & - & - & - & 31.2\s{0.7} & 50.4\s{1.2} & - & - \\
        DataDAM     & 34.7\s{0.9} & 59.4\s{0.4} & 24.2\s{0.5} & 34.4\s{0.4} & - & - & - & - & 36.4\s{0.8} & {\textbf{55.4\s{0.9}}} & - & - \\
        \midrule
        \textbf{DANCE}     & \cellcolor[HTML]{DAE8FC}{\textbf{57.2\s{0.5}}} & \cellcolor[HTML]{DAE8FC}{\textbf{80.2\s{0.7}}} & \cellcolor[HTML]{DAE8FC}{\textbf{30.6\s{0.3}}} & \cellcolor[HTML]{DAE8FC}{\textbf{57.8\s{1.1}}} & \cellcolor[HTML]{DAE8FC}{\textbf{30.6\s{0.8}}} & \cellcolor[HTML]{DAE8FC}{\textbf{52.8\s{0.7}}} & \cellcolor[HTML]{DAE8FC}{\textbf{39.4\s{0.8}}} & \cellcolor[HTML]{DAE8FC}{\textbf{60.4\s{1.1}}} & \cellcolor[HTML]{DAE8FC}{\textbf{52.0\s{0.5}}} & \cellcolor[HTML]{DAE8FC}{\textbf{77.2\s{0.3}}} & \cellcolor[HTML]{DAE8FC}{\textbf{51.8\s{1.1}}} & \cellcolor[HTML]{DAE8FC}{\textbf{78.8\s{0.7}}} \\
        \midrule
        Whole Dataset &  \multicolumn{2}{c|}{87.4\s{1.0}}  &  \multicolumn{2}{c|}{67.0\s{1.3}}  &  \multicolumn{2}{c|}{63.9\s{2.0}} &  \multicolumn{2}{c|}{66.7\s{1.1}} &  \multicolumn{2}{c|}{87.5\s{0.3}} &  \multicolumn{2}{c}{84.4\s{0.6}} \\
    \bottomrule
    \end{tabular}
    \caption{\textbf{Comparison with previous coreset selection and dataset condensation methods on high-resolution ($128\times128$) Imagenet-Subsets.} All the datasets are condensed using a 5-layer ConvNet.}
    \label{tab:res_imagenetsubsets}
\end{table*}

\section{Experiments}

\subsection{Experimental Setup}
\paragraph{Datasets.} 
We assess our method using three low-resolution datasets: Fashion-MNIST~\cite{fmnist} with a resolution of $28 \times 28$, and CIFAR-10/100~\cite{cifar} with a resolution of $32 \times 32$. For medium-resolution data, we utilize the resized TinyImageNET~\cite{tinyimage}, which has a resolution of $64 \times 64$. Furthermore, in alignment with MTT~\cite{mtt}, we employ various subsets of the high-resolution ImageNet-1K~\cite{imagenet} dataset (resolution $128 \times 128$) in our experiments. These subsets include ImageNette, ImageWoof, ImageFruit, ImageWeow, ImageSquawk, and ImageYellow. Additional details about the datasets are provided in the Appendix.

\paragraph{Network Architectures.} 
Following previous studies~\cite{mtt}, we implement the condensation process using a ConvNet~\cite{convnet}. The ConvNet we employ consists of three identical convolutional blocks, each featuring a 128-kernel $3\times 3$ convolutional layer, instance normalization, ReLU activation, and $3\times 3$ average pooling with a stride of 2. For low-resolution datasets, we use a three-layer ConvNet, while a four-layer ConvNet is utilized for TinyImageNet. To accommodate the higher resolutions of the high-resolution ImageNet-1K subsets, we employ a five-layer ConvNet.

\paragraph{Evaluation Metric.}
We utilize the test accuracy of networks trained on the condensed set $\dsyn{}$ as our primary evaluation metric. Each network is trained from scratch multiple times: 10 times for low-resolution datasets and TinyImageNet, and 3 times for the ImageNet-1K subsets. We report both the average accuracy and the standard deviation. To assess training efficiency, we consider run time per step and peak GPU memory usage as criteria, where the run time is calculated as an average over 1000 iterations.

\paragraph{Implementation Details.} For training, we employ an SGD optimizer with a learning rate of 0.01, momentum of 0.9, and weight decay of 0.0005. The expert models $\bm{\theta}_\mathsf{expert}$ are trained for 60 epochs on low-resolution datasets and TinyImageNet, and for 80 epochs on ImageNet-1K subsets. We consistently use 5 expert models for all datasets as the default setting. The number of iterations for Distribution Calibration is fixed at 1 across all datasets. During the condensing process, the SGD optimizer is set with a learning rate of 0.1 for ImageNet-1K subsets and 0.01 for other datasets, with the learning rate being scaled by the number of images per class (IPC). Following IDC~\cite{idc}, we train the networks using a sequence of color transformation, cropping, and CutMix~\cite{cutmix}. The factor parameter $l$ is set to 2 for low-resolution datasets and Tiny-ImageNet, and 3 for ImageNet-1K subsets. All synthetic data are initially generated from randomly selected real data to expedite optimization. The experiments are conducted on a GPU group comprising GTX 3090, RTX-2080, and NVIDIA-A100 GPUs.

\subsection{Comparison to State-of-The-Art Methods}
\paragraph{Baselines.}
We include a comprehensive range of methods as baselines in our study. For coreset selection methods, our choose Random Selection, Herding~\cite{core_herding}, and K-Center~\cite{core_kcenter1}. In the category of Optimization-Oriented methods, we consider DC~\cite{dc}, DSA~\cite{dsa}, IDC~\cite{idc}, DREAM~\cite{liu2023dream}, and MTT~\cite{mtt}. Additionally, for Distribution-Matching-based methods, our baselines include CAFE and CAFE+DSA~\cite{cafe}, DM~\cite{dm}, IDM~\cite{idm}, and DataDAM~\cite{sajedi2023datadam}. Further details about these baseline methods are provided in the Appendix, due to page constraints.

\begin{table}
    \centering
    \small
    \setlength\tabcolsep{3pt}
    \begin{tabular}{lc|ccc}
    \toprule
     Method & IPC & ConvNet-3 & ResNet-10 & DenseNet-121 \\
     \midrule
     \multirow{2}{*}{DSA} & 10 & 52.1\s{0.5} & 32.9\s{0.3} & 34.5\s{0.1} \\
                          & 50 & 60.6\s{0.5} & 49.7\s{0.4} & 49.1\s{0.2} \\
                          \midrule
    \multirow{2}{*}{IDC} & 10 & 67.5\s{0.5} & 63.5\s{0.1} & 61.6\s{0.6} \\
                          & 50 & 74.5\s{0.1} & \textbf{72.4\s{0.5}} & \textbf{71.8\s{0.6}} \\
        \midrule
             \multirow{2}{*}{MTT} & 10 & 56.4\s{0.7} & 34.5\s{0.8} & 41.5\s{0.5} \\
                                  & 50 & 65.9\s{0.6} & 43.2\s{0.4} & 51.9\s{0.3} \\
        \midrule
             \multirow{2}{*}{DM} & 10 & 48.9\s{0.6} & 42.3\s{0.5} & 39.0\s{0.1} \\
                                 & 50 & 63.0\s{0.4} & 58.6\s{0.3} & 57.4\s{0.3} \\
        \midrule
             \multirow{2}{*}{\textbf{DANCE}} & 10 & \textbf{70.8\s{0.2}} & \textbf{67.0\s{0.2}} & \textbf{64.5\s{0.3}} \\
                                            & 50 & \textbf{76.1\s{0.1}} & 68.0\s{0.1} & 64.8\s{0.3} \\
              \bottomrule
    \end{tabular}
    \caption{\textbf{Cross-architecture generalization performance ($\%$) on CIFAR-10.} The synthetic data is condensed using ConvNet-3 and evaluated using other architectures. The best results are in \textbf{bold}.}
    \label{tab:cross_arch}
\end{table}

\begin{figure*}[tb]
        \centering
    \includegraphics[width=\textwidth]{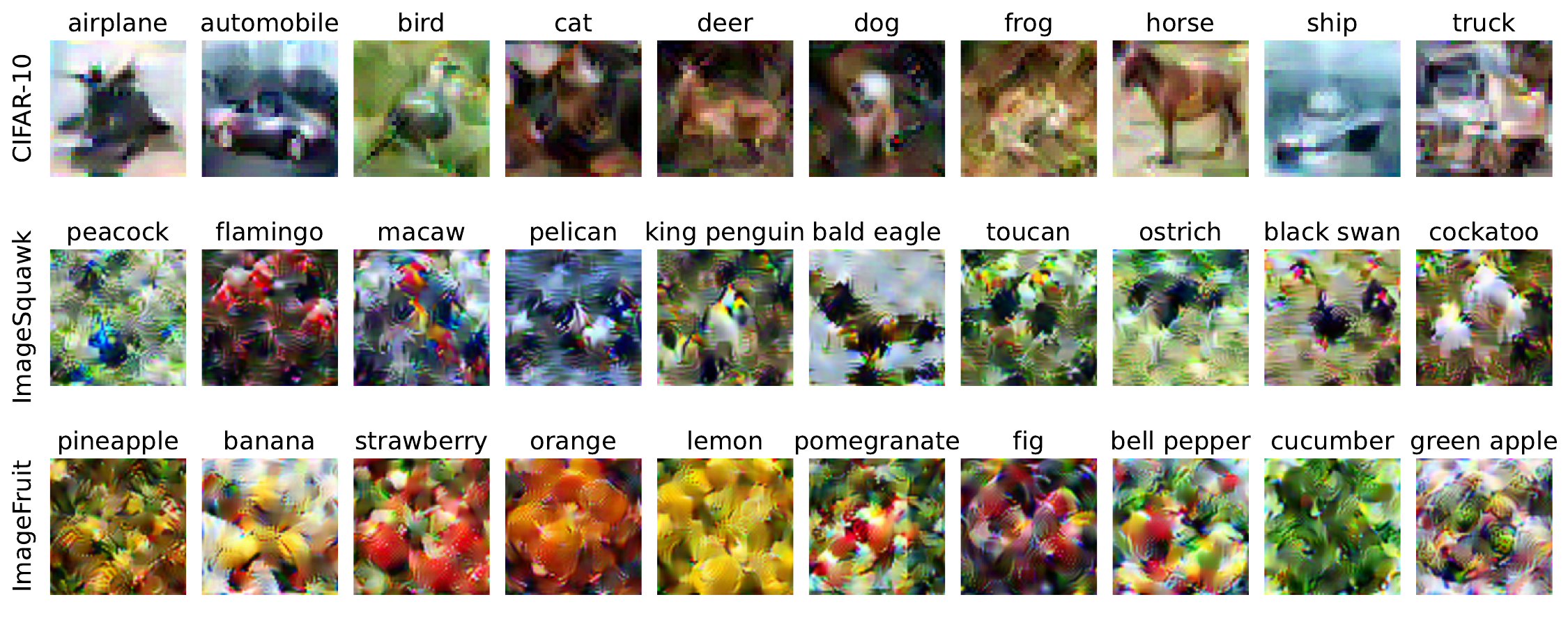}
    \caption{Example condensed images of $32\times 32$ \textbf{CIFAR-10}, $128\times 128$ \textbf{ImageSquawk}, and $128\times 128$ \textbf{ImageFruit}.}
    \label{fig:vis}
\end{figure*}

\begin{table}[tb]
    \centering
    \small
    \setlength\tabcolsep{3pt}
    \begin{tabular}{cl|rrrrrrr}
    \toprule
         & IPC & DC & DSA & DM & MTT & IDM & \textbf{DANCE} \\
         \midrule
        {\textbf{Run}} & 1 & 0.16 & 0.22 & 0.08 & 0.36 & 0.50 & 0.11 \\
        {\textbf{Time}} & 10 & 3.31 & 4.47 & 0.08 & 0.40 & 0.48 & 0.12 \\
        {\textbf{(Sec)}} & 50 & 15.74 & 20.13 & 0.08 & \textcolor{red}{\ding{55}} & 0.58 & 0.12 \\
         \midrule
        {\textbf{GPU}} & 1 & 3515 &  3513&3323  &2711  &  3223 &  2906\\
        {\textbf{Memory}} & 10 & 3621 &3639  &3455  &8049  & 3179 & 3045 \\
        {\textbf{(MB)}} & 50 & 4527 &4539  & 3605  & \textcolor{red}{\ding{55}} & 4027 & 3549 \\
         \bottomrule
    \end{tabular}
        \caption{\textbf{Time and GPU memeory cost comparison of SOTA datasets condensation methods.} Run Time: the time for a single iteration. GPU memory: the peak memory usage during condensing. Both run time and GPU memory is averaged over 1000 iterations. All experiments are conducted on CIFAR-10 with a single NVIDIA-A100 GPU. ``\textcolor{red}{\ding{55}}'' denotes out-of-memory issue.}
    \label{tab:timeandmemory}
\end{table}

\paragraph{Performance Comparison.}
Tab.~\ref{tab:res_lowresolution} and Tab.~\ref{tab:res_imagenetsubsets} present the comparison of our method with coreset selection methods and dataset condensation/distillation methods. The proposed method, {DANCE}, demonstrates remarkable performance across various datasets and resolutions. On low-resolution datasets such as Fashion-MNIST, CIFAR-10, CIFAR-100, and the medium-resolution dataset TinyImageNet, {DANCE} consistently outperforms or rivals leading methods in different IPC (images per class) settings. For instance, on Fashion-MNIST, it achieves the highest test accuracy of 81.5$\%$ with a single IPC. On CIFAR-10 and CIFAR-100, {DANCE} sets new benchmarks with 70.8$\%$ and 52.8$\%$ accuracy respectively at 50 IPC, even surpassing the SOTA optimization-oriented methods DREAM and IDC. Particularly notable is its performance on TinyImageNet, where it attains an accuracy of 11.6$\%$ at 1 IPC and 26.4$\%$ at 10 IPC, significantly ahead of the next best method, IDM. For the high-resolution ImageNet-1K subsets, DANCE still yield SOTA results across various scenarios. Remarkably, across all ImageNet-1K subsets with 10 images per class, our DANCE brings over 10$\%$ accuracy increase compared to the second best results, showcasing its efficacy in handling diverse image complexities. These results show the superiority of DANCE in dataset condensation tasks, especially considering the wide margin by which it leads in many categories. Overall, {DANCE} not only establishes new standards in dataset condensation but also demonstrates its robustness across varying resolutions and dataset complexities.

\paragraph{Cross-Architecture Evaluation.}
We also evaluated the performance of our condensed set across various architectures, as detailed in Tab.~\ref{tab:cross_arch}. The results demonstrate that DANCE excels not only on the architecture employed during the condensation process but also exhibits impressive generalization capabilities on a range of unseen architectures

\paragraph{Training Efficiency Evaluation.}
In the context of dataset condensation, it is of great importance to consider the resource and time costs, as extensively discussed in previous studies~\cite{sajedi2023datadam,modelaugmentation}. Some of the methods entail significantly higher time costs in comparison to the time required for training the entire dataset, rendering them less than optimal in balancing effectiveness and efficiency. Our evaluation encompasses both time and peak GPU memory costs incurred during the condensation process for various baseline methods and DANCE. As presented in Tab.~\ref{tab:timeandmemory}, DANCE exhibits remarkable efficiency compared to optimization-oriented methods such as DC~\cite{dc}, DSA~\cite{dsa}, and MTT~\cite{mtt}. Much like DM~\cite{dm}, our method demonstrates scalability across different IPCs. However, IDM, being rooted in the DM-based approach, displays higher time costs when contrasted with both DM~\cite{dm} and DANCE.

\subsection{Ablation Studies}\label{sec:ablation}
\paragraph{Effectiveness of Each Module.}
We evaluate three primal Modules of our method, namely Pseudo Long-Term Distribution Alignment (Sec.~\ref{sec:inner}), Distribution Calibration (Sec.~\ref{sec:inter}), and Factoring \& Up-sampling technique (Sec.~\ref{sec:train_algor}). As shown in Tab.~\ref{tab:ablation_eachcomponent}, both the proposed PLTDA and Distribution Calibration bring significant improvement across various datasets. The most significant improvement is observed when all three modules are included. The results highlight the effectiveness of the three modules, demonstrating their collective importance in enhancing the DANCE framework's performance across different datasets.

\begin{table}[tb]
    \centering
    \small
    \setlength\tabcolsep{1.5pt}
    \begin{tabular}{ccc|cc|cc}
    \toprule
    \multirow{2}{*}{Fac.} & \multirow{2}{*}{PLTDA} & \multirow{2}{*}{Dist. Calib.} &  \multicolumn{2}{c|}{CIFAR-10} &  \multicolumn{2}{c}{CIFAR-100} \\
        & &  & 10 & 50 & 10 & 50\\
        \midrule
        - & \ding{52} & \ding{52} & 56.1\s{0.2} & 71.4\s{0.4} & 40.3\s{0.2} & 50.6\s{0.1}\\
        \ding{52} & - & \ding{52} & 64.8\s{0.1} & 68.2\s{0.1} & 37.2\s{0.1} & 45.6\s{0.2}\\
        \ding{52} & \ding{52} & - & 65.6\s{0.3} & 69.8\s{0.2} & 43.5\s{0.3} & 47.5\s{0.2}\\
        \ding{52} & \ding{52} & \ding{52} & \textbf{70.8\s{0.2}} & \textbf{76.1\s{0.1}} & \textbf{49.8\s{0.1}} & \textbf{52.8\s{0.1}} \\
        \bottomrule
    \end{tabular}
    \caption{{\textbf{Ablation study on three main modules of DANCE.} ``\ding{52}'' denotes the module is included, and ``-'' ortherwise. ``Fac.'' denotes the Factoring technique. ``Dist. Calib.'' denotes the module of Distribution Calibration.}}
    \label{tab:ablation_eachcomponent}
\end{table}

\paragraph{Impact on the Number of Expert Models.}
The expert models $\bm{\theta}_\mathsf{expert}$ are integral to both the PLTDA and Distribution Calibration modules within DANCE. To ascertain their impact, we investigated how the number of expert models (NEM) affects DANCE's performance. As Tab.~\ref{tab:ablation_expertmodels} illustrates, there is a noticeable increase in DANCE's performance with the rise in NEM. Notably, even with just a single expert model, DANCE achieves competitive results, scoring 69.2$\%$ on CIFAR-10 with 10 images per class. This underscores DANCE's ability to efficiently leverage the pre-trained knowledge embedded in expert models.
\begin{table}[tb]
    \centering
    \small
    \setlength\tabcolsep{3pt}
    \begin{tabular}{c|cccccccc}
    \toprule
       NEM  & 1 & 2 & 3 & 4 & 5 & 10 & 15 & 20 \\
       \midrule
       Acc. & 69.2 & 70.1 & 70.2 & 70.2 & 70.8 & 71.2 & 71.1 & 71.1 \\
         \bottomrule
    \end{tabular}
        \caption{\textbf{Ablation on the number of expert models (NEM).} The evaluation is conducted on CIFAR-10 with 10 images per class.}
    \label{tab:ablation_expertmodels}
    \vspace{-5pt}
\end{table}

\subsection{Visualization Results}
In Fig.~\ref{fig:vis}, we present the synthetic images condensed by DANCE, showcasing distinct characteristics across different datasets. For the low-resolution dataset CIFAR-10, the condensed images are quite discernible, with each clearly representing its respective class. In contrast, the condensed images from the high-resolution ImageNet-1K subsets appear more abstract and outlined. Unlike the images produced by DM~\cite{dm}, which feature class-independent textures, our synthetic images encapsulate richer information pertinent to classification tasks. Additional visualizations are available in the Appendix due to page limitations.

\section{Conclusion}\label{sec:conclusion}
In this study, we introduce a novel framework called \underline{\textbf{D}}ual-view distribution \underline{\textbf{A}}lignment for dataset \underline{\textbf{C}}ondensation (DANCE), which enhances the Distribution Matching (DM) method by focusing on both inner- and inter-class views. DANCE consists of two meticulously designed modules: Pseudo Long-Term Distribution Alignment (PLTDA) for inner-class view and Distribution Calibration for inter-class view. PLTDA ensures that the data condensed by DANCE effectively represents its class throughout the entire training process while eliminating the need for persistent training. In contrast, Distribution Calibration maintains the synthetic data within its respective class region. Extensive experimental results on various datasets show that DANCE consistently surpasses state-of-the-art methods while requiring less computational costs. This makes DANCE highly suitable for practical and complex scenarios.

\section*{Acknowledgments}
This work was partially supported by grants from the Pioneer R\&D Program of Zhejiang Province (2024C01024).

\bibliographystyle{named}

\bibliography{ijcai24}

\appendix
\onecolumn

\section*{Related Works}

\subsection*{Optimization-Oriented Methods}~
Optimization-oriented methods learn the synthetic dataset via a bi-level optimization or a meta-learning framework~\cite{liu2021investigating,mtt,dc,idc,dsa,modelaugmentation,du2023minimizing,dd_wang}.
The pioneering work~\cite{dd_wang} poses a strong assumption that a model trained on the synthetic dataset should be identical to that trained on the real dataset. 
Due to vast parameter space and convergence challenges for matching the converged models, subsequent works adopt a more stringent assumption that the two models trained on synthetic dataset and real dataset should follow a similar optimization path, which can be realized by \textit{performance matching} or \textit{parameter matching}.

\myPara{Performance Matching.}~In performance matching, the synthetic dataset is optimized to ensure the model trained on it achieves the lowest loss on the real dataset~\cite{dd_wang}, in which way the performance of models could be matched. Further, the kernel ridge regression (KRR)-based methods are proposed to mitigate the inefficiency of the meta-gradient back-propagation~\cite{nguyen2020dataset,KIP,zhou2022dataset,loo2022efficient,loo2023dataset}. With KRR, the synthetic dataset can be updated by back-propagating meta-gradient through the kernel function~\cite{nguyen2020dataset}. Following the KRR stream, KIP~\cite{KIP} employs infinite-width neural networks as its kernel function, thereby forging a significant link between KRR and the field of deep learning.

\myPara{Parameter Matching.}~
The concept of parameter matching in dataset condensation was initially introduced by DC~\cite{dc}, and has since been expanded upon in various subsequent studies~\cite{dsa,mtt,idc}. The fundamental principle of this approach is to align the parameters induced by real and synthetic datasets. In the initial study~\cite{dc}, the focus was on minimizing the difference between gradients derived from synthetic and real datasets with respect to the model. Subsequently, DSA~\cite{dsa} enhanced this by incorporating differentiable siamese augmentation prior to feeding examples into the model, thereby increasing the synthetic dataset's informativeness. Meanwhile, MTT~\cite{mtt} addressed the potential error accumulation in single-step gradients by introducing a multi-step parameter matching method. This method iteratively updates synthetic data to align the model's training trajectory on synthetic dataset with that on real dataset.

Concurrently, some studies have concentrated on refining the model update process during condensation. DC~\cite{dc} utilized the synthetic dataset for updating the network, which risked early-stage training over-fitting. IDC~\cite{idc} countered this by updating the model using real datasets, reducing over-fitting risks due to the larger size of real datasets. IDC~\cite{idc} also introduced a factoring technique to enhance the synthetic dataset's richness through factoring and up-sampling. Despite their effectiveness, these parameter matching methods are resource-intensive, requiring numerous differently initialized networks to update the synthetic dataset. To expedite this process, \cite{modelaugmentation} proposed model augmentation, introducing Gaussian perturbation to early-stage models to reduce the time and storage requirements for dataset condensation.

\subsection*{Distribution-Matching-based Methods}~
Distribution-matching-based methods aim to create a synthetic dataset that keeps similar representation distribution of the real dataset~\cite{cafe,dm,idm,sajedi2023datadam}. These methods differ from optimization-oriented ones in that they bypass the need for bi-level optimization or meta-gradient use in dataset condensation, significantly cutting down on time and memory costs. DM~\cite{dm}, for instance, simplifies the process by aligning the representation embeddings of real and synthetic examples, and omits the model updating step, as this has minimal impact on synthetic example performance. CAFE~\cite{cafe} takes this a step further by aligning embeddings not just in the last layer but also in earlier layers, enhancing the synthetic dataset's discriminative qualities through a discriminant loss term. IDM~\cite{idm} introduces a ``partitioning and expansion'' technique to boost the number of representations drawn from the synthetic dataset, addressing the class misalignment issue found in DM~\cite{dm} by using a trained model with a cross-entropy regularization loss. DataDAM~\cite{sajedi2023datadam} adds spatial attention matching to improve the synthetic set's performance.

\subsection*{Coreset Selection}~
Coreset selection methods, as opposed to synthesizing data, focus on choosing a subset from the entire training set based on specific criteria, as seen in approaches like Herding~\cite{core_herding}, K-center~\cite{core_kcenter1,core_kcenter2}, and others. Herding~\cite{core_herding}, for example, picks samples near the centers of their respective classes, while K-center~\cite{core_kcenter1} aims to minimize the maximum distance between chosen samples and their nearest class center by selecting several center points within a class. However, the effectiveness of these coresets is not always assured due to the heuristic nature of the selection criteria. Additionally, the potential of coresets is limited by the quality of the original training examples, posing a challenge to their use in reducing data requirements.

\section*{Details of Datasets}


\subsection*{Low-Resolution Datasets}
\begin{itemize}
    \item \textbf{Fashion-MNIST}~\cite{fmnist}, a widely recognized dataset, is frequently employed for assessing machine learning models. It encompasses 60,000 images for training and 10,000 for testing, each rendered in gray-scale and measuring $28\times28$ pixels. This dataset features an array of 10 distinct fashion categories, encompassing various items such as T-shirts, dresses, and shoes.
    \item \textbf{CIFAR-10/100}~\cite{cifar} are extensively utilized benchmark datasets in the realm of object recognition and classification. CIFAR-10 is composed of 60,000 color images, divided into 50,000 for training and 10,000 for testing, spanning 10 diverse object classes, such as cars, birds, and cats. In contrast, CIFAR-100 encompasses a broader range of 100 object classes, allocating 600 images to each class. Each image in both datasets is $32\times32$ pixels, rendering them ideal for testing and evaluating algorithms in image classification and object recognition tasks.
\end{itemize}

\subsection*{Medium-Resolution Datasets}
\begin{itemize}
    \item \textbf{TinyImageNet}~\cite{tinyimage}, a streamlined version of the larger ImageNet dataset~\cite{imagenet}, is a widely recognized benchmark in the field of image recognition and classification. This dataset is composed of 100,000 color images for training, alongside 10,000 images reserved for validation and another 10,000 for testing. Each image in TinyImageNet is $64\times64$ pixels, providing a more compact yet challenging dataset for algorithm evaluation. It covers 200 diverse object classes, ranging from everyday items to various animals and scenes, offering a rich and varied dataset for tasks in image classification and object recognition.
\end{itemize}

\subsection*{High-Resolution Datasets}
\begin{itemize}
    \item \textbf{ImageNet Subsets} Following MTT~\cite{mtt}, we adopt six subsets of ImageNet~\cite{imagenet} as high-resolution ($128\times128$) datasets to evaluate our method, including ImageNette (assorted objects) and ImageWoof (dog breeds), ImageFruit (fruits), ImageMeow (cats), ImageSquawk (birds), and ImageYellow (yellow-ish things). Each of these subsets consists of 10 distinct classes.
\end{itemize}

\section*{Details of Baselines}

\subsection*{Coreset-Selection}
\begin{itemize}
    \item \textbf{Random}: randomly select a subset of the original dataset for training.
    \item \textbf{Herding}: selecting data points that are close to the class centres~\cite{core_herding}.
    \item \textbf{K-center}: selecting the subset using K-center algorithm, which iteratively selects centers and including points that are closest to these centers~\cite{core_kcenter1,core_kcenter2}.
\end{itemize}

\subsection*{Dataset-Condensation}
\noindent\textbf{Optimization-Oriented.}
\begin{itemize}
    \item \textbf{DC}: iteratively updating the network on the synthetic dataset and matching the gradient induced by the real and synthetic images~\cite{dc}.
    \item \textbf{DSA}: applying a differentiable siamese augmentation to images before feed them to the network~\cite{dsa}.
    \item \textbf{IDC}: using a factoring technique that split one image into several lower-resolution ones. Besides, IDC update the network on the original real set instead of the condensed set~\cite{idc}.
    \item \textbf{DREAM}: combining IDC with a distribution-aware data sampler, which emploit K-Means method to select a more evenly-distributed real data to guild the update of synthetic dataset~\cite{liu2023dream}.
    \item \textbf{MTT}: matching the training trajectories induced by real and synthetic datasets~\cite{mtt}.
\end{itemize}
\noindent\textbf{Distribution-Matching-based.}
\begin{itemize}
    \item \textbf{CAFE}: aligning the feature embedding of the real and synthetic images in a layer-wise manner. Moreover, CAFE utilizes a discriminant loss to enhance the discriminative properties of the synthetic dataset~\cite{cafe}.
    \item \textbf{CAFE+DSA}: additionally combining DSA strategy to images compared to CAFE~\cite{cafe}.
    \item \textbf{DM}: aligning the feature embedding of the real and synthetic datasets~\cite{dm}.
    \item \textbf{IDM}: introducing the partitioning and expansion technique and a distribution regularization to improve the original DM~\cite{idm}.
    \item \textbf{DataDAM}: except for the loss of DM~\cite{dm}, DataDAM also incorporate the attention matching loss term to utilize the spatial attention information~\cite{sajedi2023datadam}.
\end{itemize}

\section*{More Visualization Results}

\begin{figure*}[tb]
        \centering
    \includegraphics[width=\textwidth]{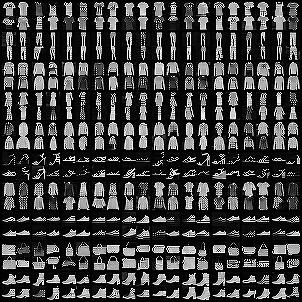}
    \caption{Condensed images of \textbf{Fashion-MNIST} with 10 images per class.}
\end{figure*}

\begin{figure*}[tb]
        \centering
    \includegraphics[width=\textwidth]{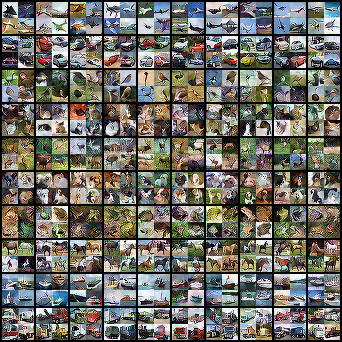}
    \caption{Condensed images of \textbf{CIFAR-10} with 10 images per class.}
\end{figure*}

\begin{figure*}[tb]
        \centering
    \includegraphics[width=\textwidth]{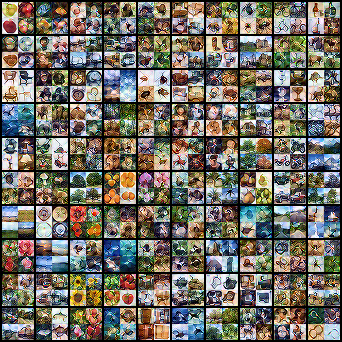}
    \caption{Condensed images of \textbf{CIFAR-100} with 1 image per class.}
\end{figure*}

\begin{figure*}[tb]
        \centering
    \includegraphics[width=\textwidth]{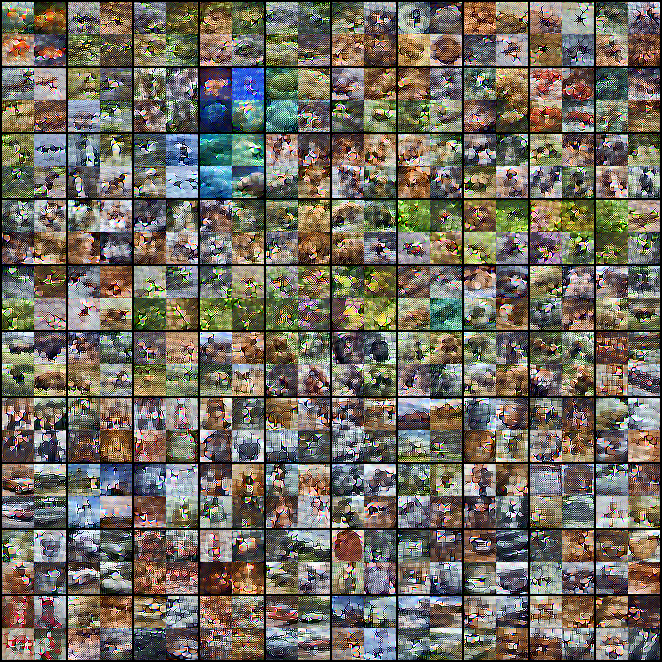}
    \caption{Condensed images of \textbf{TinyImageNet} with 1 images per class (part 1: class 0 to 99).}
\end{figure*}
\begin{figure*}[tb]
        \centering
    \includegraphics[width=\textwidth]{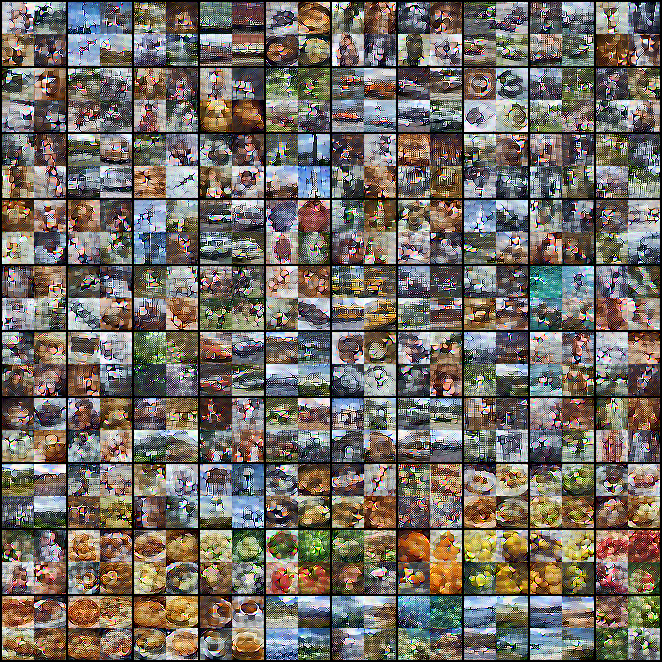}
    \caption{Condensed images of \textbf{TinyImageNet} with 1 images per class (part 2: class 100 to 199).}
\end{figure*}

\begin{figure*}[tb]
        \centering
    \includegraphics[width=\textwidth]{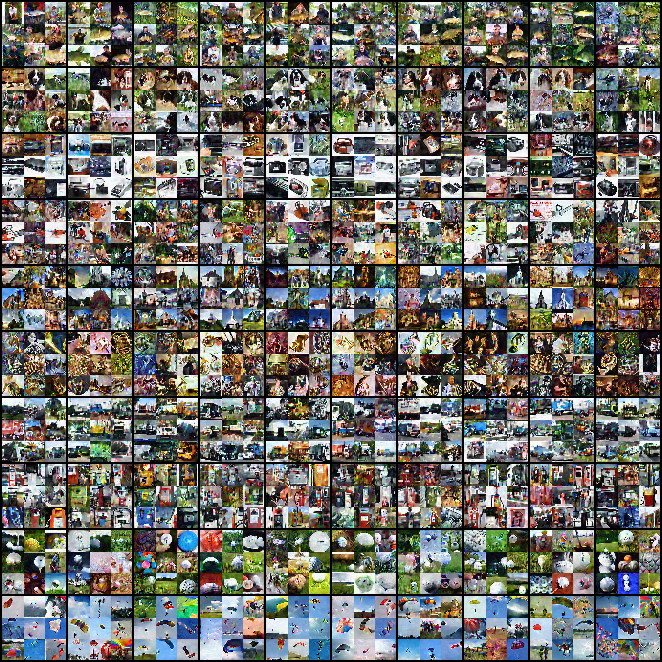}
    \caption{Condensed images of \textbf{ImageNette} with 10 images per class.}
\end{figure*}

\begin{figure*}[tb]
        \centering
    \includegraphics[width=\textwidth]{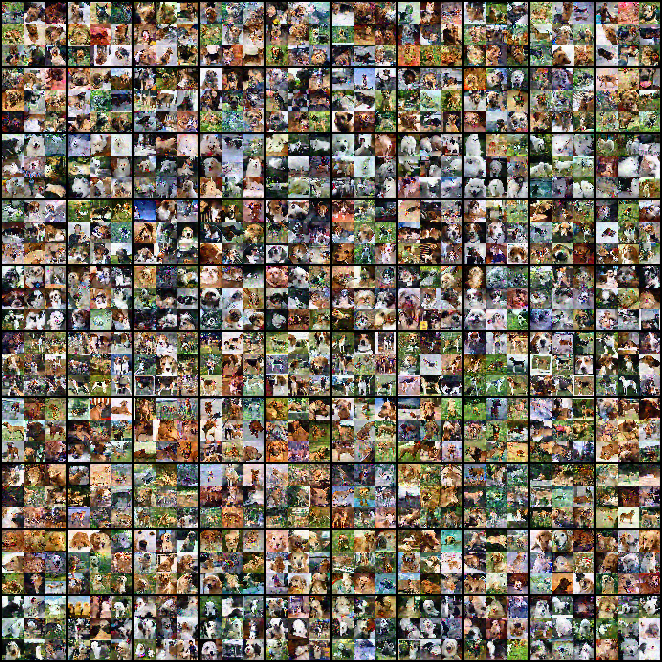}
    \caption{Condensed images of \textbf{ImageWoof} with 10 images per class.}
\end{figure*}

\begin{figure*}[tb]
        \centering
    \includegraphics[width=\textwidth]{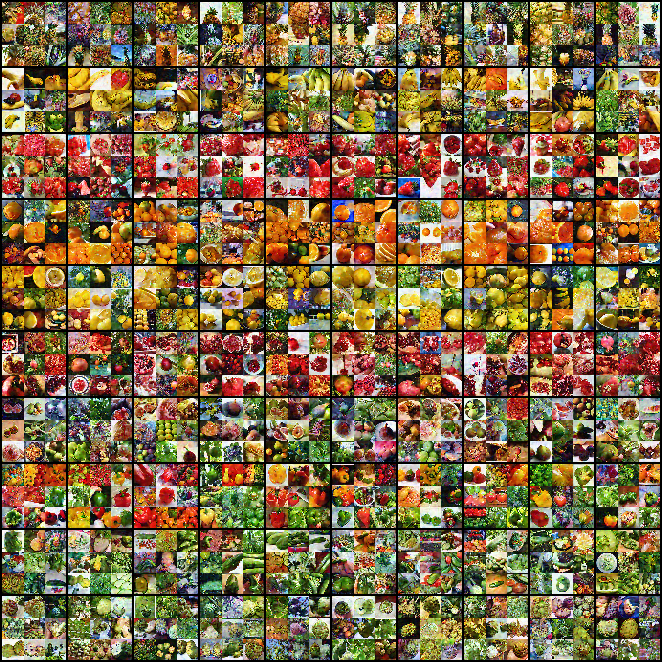}
    \caption{Condensed images of \textbf{ImageFruit} with 10 images per class.}
\end{figure*}

\begin{figure*}[tb]
        \centering
    \includegraphics[width=\textwidth]{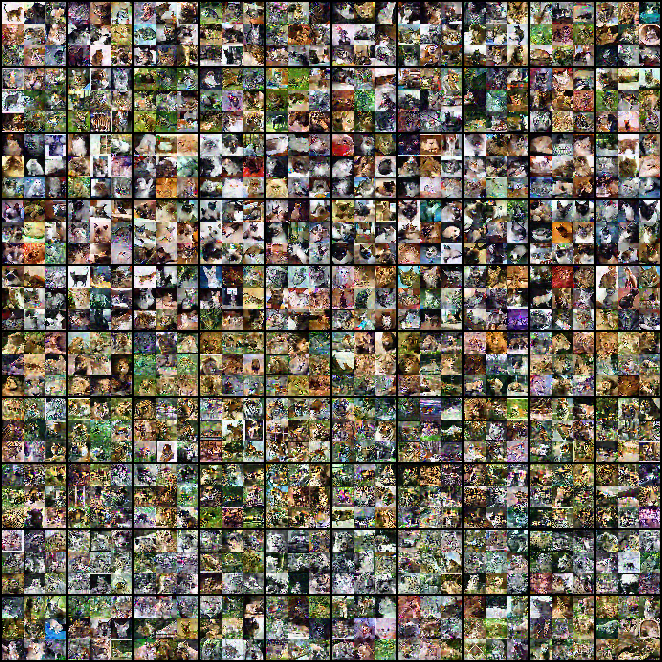}
    \caption{Condensed images of \textbf{ImageMeow} with 10 images per class.}
\end{figure*}

\begin{figure*}[tb]
        \centering
    \includegraphics[width=\textwidth]{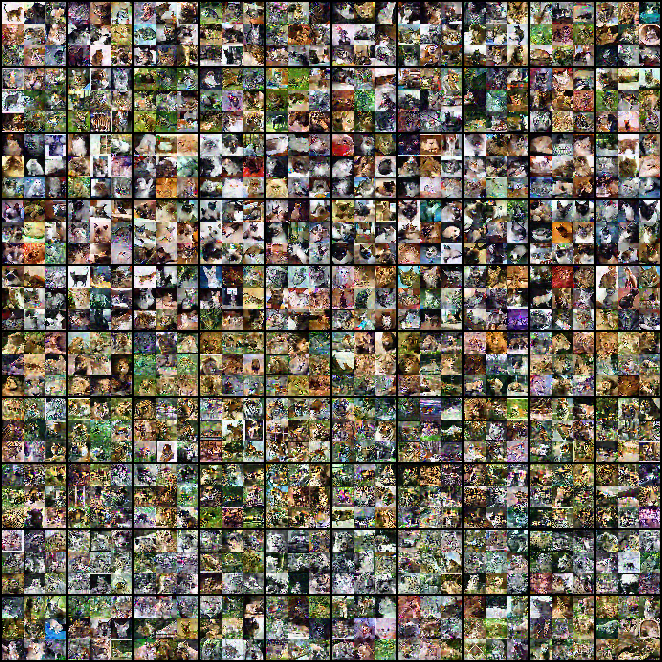}
    \caption{Condensed images of \textbf{ImageSquawk} with 10 images per class.}
\end{figure*}

\begin{figure*}[tb]
        \centering
    \includegraphics[width=\textwidth]{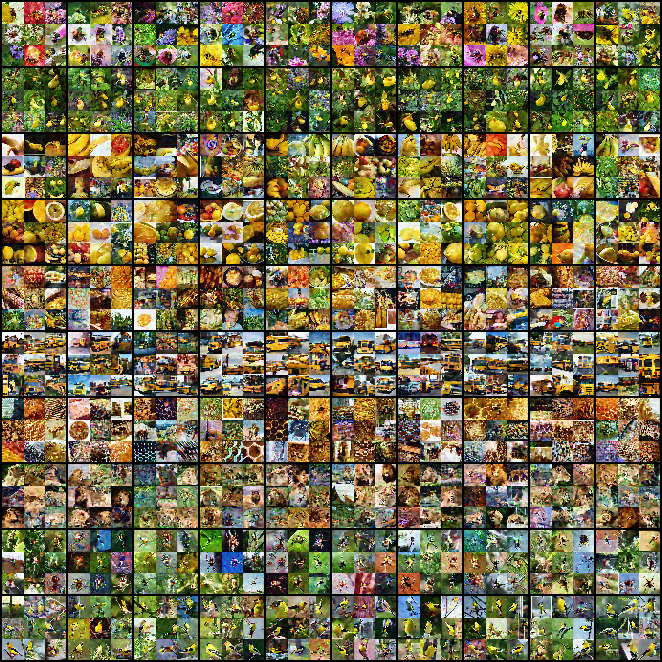}
    \caption{Condensed images of \textbf{ImageYellow} with 10 images per class.}
\end{figure*}




\end{document}